\documentclass[conference]{IEEEtran}
\IEEEoverridecommandlockouts

\usepackage{mathptmx}  
\usepackage{helvet}    

\usepackage{amsmath,amssymb,amsfonts}

\usepackage{algorithm}
\usepackage{algorithmic}

\usepackage{graphicx}
\usepackage{subfig}
\usepackage{makecell}
\usepackage{multirow}
\usepackage{float}

\usepackage{textcomp}
\usepackage{xcolor}
\usepackage{enumitem}

\usepackage[hyphens]{url}

\usepackage{cite}
\usepackage{hyperref} 

\def\BibTeX{{\rm B\kern-.05em{\sc i\kern-.025em b}\kern-.08em
    T\kern-.1667em\lower.7ex\hbox{E}\kern-.125emX}}

\usepackage{fancyhdr}
\usepackage{tikz}
\newcommand{\copyrightfooter}{
    \begin{tikzpicture}[remember picture,overlay]
        \node[anchor=south, yshift=0.5cm] at (current page.south) {  
            \begin{minipage}{\textwidth}
                \centering
                \fbox{\parbox{\dimexpr\textwidth-0.2cm\relax}{ 
                \small \textcopyright 2025 IEEE. The final published paper is copyrighted by IEEE and should be cited as:
                Bishal K C, Amr Hilal, Pawan Thapa, “Anomaly Detection in Electric Vehicle Charging Stations Using Federated Learning,” IEEE Global Conference on Artificial Intelligence \& Internet of Things (GCAIoT), 2025, pp. 1–7. Personal use of this material may be permitted and permission from IEEE must be obtained for all other uses.
                }}
            \end{minipage}
        };
    \end{tikzpicture}
}

\fancypagestyle{firstpage}{
    \fancyhf{}
    \fancyhead[L]{\small {THIS PAPER HAS BEEN ACCEPTED IN \underline{\textit{IEEE GCAIoT 2025}} CONFERENCE AND FINAL VERSION IS SUBMITTED.}}
    \fancyfoot[C]{\copyrightfooter}
}

\begin{document}

\title{Anomaly Detection in Electric Vehicle Charging Stations Using Federated Learning \\
}

\author{\IEEEauthorblockN{ Bishal K C}
\IEEEauthorblockA{\textit{Department of Computer Science} \\
\textit{Tennessee Technological University}\\
Cookeville, TN USA \\
bkc42@tntech.edu}
\and
\IEEEauthorblockN{Amr Hilal}
\IEEEauthorblockA{\textit{Department of Computer Science} \\
\textit{Tennessee Technological University}\\
Cookeville, TN USA \\
ahilal@tntech.edu}
\and
\IEEEauthorblockN{Pawan Thapa}
\IEEEauthorblockA{\textit{Department of Engineering Technology} \\
\textit{University of Toledo}\\
Toledo, OH USA \\
pawan.thapa@rockets.utoledo.edu}
}

\maketitle
\thispagestyle{firstpage}

\begin{abstract}

Federated Learning (FL) is a decentralized training framework widely used in IoT ecosystems that preserves privacy by keeping raw data local, making it ideal for IoT-enabled cyber-physical systems with sensing and communication like Smart Grids (SGs), Connected and Automated Vehicles (CAV), and Electric Vehicle Charging Stations (EVCS). With the rapid expansion of electric vehicle infrastructure, securing these IoT-based charging stations against cyber threats has become critical. Centralized Intrusion Detection Systems (IDS) raise privacy concerns due to sensitive network and user data, making FL a promising alternative. However, current FL-based IDS evaluations overlook practical challenges such as system heterogeneity and non-IID data. To address these challenges, we conducted experiments to evaluate the performance of federated learning for anomaly detection in EV charging stations under system and data heterogeneity. We used FedAvg and FedAvgM, widely studied optimization approaches, to analyze their effectiveness in anomaly detection. Under IID settings, FedAvg achieves superior performance to centralized models using the same neural network. However, performance degrades with non-IID data and system heterogeneity. FedAvgM consistently outperforms FedAvg in heterogeneous settings, showing better convergence and higher anomaly detection accuracy. Our results demonstrate that FL can handle heterogeneity in IoT-based EVCS without significant performance loss, with FedAvgM as a promising solution for robust, privacy-preserving EVCS security.
\end{abstract}


\section{Introduction}
The number of electric vehicles (EVs) is rapidly growing as people are shifting towards renewable energy sources due to the adverse harmful effects associated with burning fossil fuels. According to the latest forecast by Gartner Inc.~\cite{gartner2024}, the number of electric vehicles is expected to grow by 33\% in 2025, reaching 85 million EVs worldwide by the end of the year. Additionally, 10.4 million EVs are projected to be in use in North America alone by 2025. However, this rapid growth introduces a significantly larger attack surface due to the increased interconnectivity between networks, such as charging stations, smart grids, and Vehicle Multimedia Systems, which makes them more vulnerable to cyberattacks. These charging stations function as interconnected IoT devices, exchanging operational data with vehicles, grid operators, and cloud-based energy management platforms, which makes them part of a larger IoT-enabled smart transportation ecosystem. As a result, there is a high risk of cybersecurity threats targeting EVCS, smart grid systems, vehicles, and even end users. To mitigate such threats, a robust Intrusion Detection System (IDS) is required to detect attacks in advance and prevent them. \\ 
\indent Existing IDS approaches, such as ensemble-based IDS and deep neural network-based IDS, require centralized data collection for model training. However, EV user data contains sensitive personal information and collecting such data raises significant privacy concerns. In the IoT domain, Federated Learning (FL) has emerged as an effective privacy-preserving technique that enables decentralized training across distributed, resource-constrained devices, making it highly suitable for EVCS networks. 
Several studies have been conducted. For example, Purohit et al.~\cite{purohit2024} proposed FL-EVCS, a FL-based anomaly detection system for EVCS, which enhances cybersecurity by preserving data privacy while improving detection accuracy. Their study demonstrated high performance, achieving around 97\% accuracy on the CICEVSE2024 dataset, outperforming a traditional ML-based Anomalies Detection System. Similarly, Bamini et al.~\cite{bamini2024} used the Enhanced Dynamic Threshold Whale Optimization-based Feature Selection (EDTWFS) method to select features for intrusion detection in EVCS and implemented the Federated Averaging Learning Classifier (FALC) for intrusion detection in EVCS, which preserves data privacy while enhancing security. Similarly, Mothukuri et al.~\cite{mothukuri2022} introduced an FL-based anomaly detection for IoT security attacks utilizing Gated Recurrent Units (GRUs) to detect anomalies in network traffic. 

These prior works primarily focus on scenarios with IID data distributions and assume uniform system capabilities across participating clients. These assumptions are rarely valid in real-world EVCS environments, where system heterogeneity (such as hardware, latency, and energy differences) and non-IID data distributions are common. Similarly, for instance, Das et al.~\cite{das2020electric} highlight how the diverse mix of charging technologies, ranging from simple passive loads to active, bidirectional (V2G) assets, creates a fundamentally heterogeneous environment for grid operators. Similarly, Ma et al.~\cite{ma2024exploring} examine the broad range of communication standards used in EV charging, from short-range protocols like Bluetooth for authentication to vehicle-to-everything (V2X) for dynamic control, resulting in a highly heterogeneous communication infrastructure. Addressing these challenges is crucial for deploying robust and practical FL-based IDS in realistic EVCS infrastructures.
In this research, the following research questions are addressed:
\begin{itemize}
    \item How can we simulate EV charging station environments with data and system heterogeneity in a FL setting?
    \item How does the performance of FL compare to its centralized counterpart using the same model architecture in detecting cyber attacks while preserving privacy?
\end{itemize}
Addressing the above-mentioned research questions is critical for ensuring the security and reliability of EV charging networks. The advantage of implementing FL is that it ensures that EVCS and EV user data remain private and are not exposed to centralized databases, reducing possible data leakage risks. Similarly, it will also help mitigate potential financial, operational, and reputational damages caused by security breaches by detecting cyber threats early. Moreover, the decentralized nature of FL does not require extensive cloud infrastructure, thus enabling scalable anomaly detection across a vast network of EV charging stations. In short, this study provides new insights into how FL can be optimized for anomaly detection in critical infrastructure while maintaining high detection accuracy and minimizing communication costs.

\section{Literature Review}
Several studies have been conducted to identify anomalies in EVCS using different machine learning and FL-based detection methods. Basnet et al.~\cite{basnet2020deep} applied Deep Neural Networks (DNNs) and Long Short-Term Memory (LSTM) algorithms to detect anomalies, achieving over 99\% accuracy. Nguyen et al.~\cite{nguyen2019} developed D\"{I}oT, a federated self-learning anomaly detection system that builds device type-specific behaviour models. It detects anomalies in IoT communication, achieving high accuracy with zero false alarms using symbol-based packet analysis and GRUs.
Similarly, Alsulaimawi~\cite{alsulaimawi2024federated} integrates gradient-based anomaly detection with autoencoder-driven data reconstruction to effectively mitigate poisoned data. This method is validated on the MNIST and CIFAR-10 datasets and improves anomaly detection accuracy by 15\% over existing solutions while maintaining a low false positive rate, highlighting its potential for securing FL deployments in critical sectors such as healthcare and finance. Dong et al.~\cite{dong2025} proposed FADngs, a federated anomaly detection method that shares noisy density functions to align anomaly definitions across clients while preserving privacy. They utilize contrastive learning for enhanced feature representation and ensemble distillation to enhance global model performance. However, only a few of the empirical studies have been conducted on anomaly detection in EVCS. Likewise, Purohit et al.~\cite{purohit2024} proposed FL-EVCS, a FL-based anomaly detection system for EVCS, while Bamini et al.~\cite{bamini2024} applied feature selection with Enhanced Dynamic Threshold Whale Optimization-based Feature Selection (EDTWFS) and used the Federated Averaging Learning Classifier (FALC) to enhance security and privacy.

Despite the existing work in literature, these studies did not consider system heterogeneity, a critical factor in real-world EVCS environments. Variations in hardware capabilities, network latency, local dataset sizes and energy constraints can significantly affect model performance in FL-based anomaly detection. For instance, different charging standards such as the Combined Charging System (CCS) support multiple connector types to accommodate regional hardware differences~\cite{ma2024exploring}. Network variability also impacts EVCS performance as highlighted by SMERC’s work on EV-grid communication technologies~\cite{smerc}. Furthermore, energy constraints, especially with Vehicle-to-Grid (V2G) technology also impact resource availability and model training. To address these challenges, this study evaluates Federated Averaging with Momentum (FedAvgM) for anomaly detection in EVCS using a non-IID dataset. The research specifically analyzes the effects of system heterogeneity that arises from hardware \& network data and demonstrates that FedAvgM effectively improves convergence and performance under such conditions, filling a critical gap in prior works.

\section{Dataset}
For this experiment, the latest published open-source dataset i.e. CIC EV Charger Attack Dataset 2024 (CICEVSE2024) ~\cite{buedi2024enhancing} was used which is developed by a researcher at the Canadian Institute of Cybersecurity. This dataset ~\cite{CICEVSE2024} is a multi-dimensional labelled dataset containing benign and attack scenarios. The attack-labelled dataset comprises network \& host attacks on the Electric Vehicle Supply Equipment (EVSE) charger in both idle \& charging states. Moreover, network attacks consist of various reconnaissance \& Denial-of-Service (DoS) attacks, while host attacks include backdoors \& cryptojacking. The testbed setup consists of an operational Level 2 charging station (EVSE-A), a Raspberry Pi, and communication equipment. Raspberry Pis are used for the Electric Vehicle Communication Controller (EVCC), another charging station (EVSE-B), a Power Monitor and the local Charging Station Monitoring System (CSMS). EVSE-A communicates with a remote CSMS via the OCPP protocol while EVSE-B interacts with the EVCC using ISO 15118 and the local CSMS via OCPP. This dataset ~\cite{CICEVSE2024} was selected for its domain specificity, as it captures real EVCS attack scenarios and operational behaviours, including protocol-level interactions like OCPP and ISO 15118. This makes it more suitable than generic intrusion datasets for evaluating anomaly detection in EVCS environments. While the proposed model could be applied to other datasets, its demonstrated performance here highlights its effectiveness in EVCS-specific contexts. For the experiment, HPC and kernel events from the EVSE-B dataset were utilized to train the model to detect attacks.

\section{Preliminaries}
Federated Learning (FL) is a decentralized collaborative machine learning technique that allows multiple clients to collaboratively train a shared machine learning model, keeping their data private rather than centrally stored. This approach enhances data privacy and security by ensuring that sensitive local datasets remain on-device. In FL, a widely used optimization algorithm, Federated Averaging (FedAvg) operates in iterative communication rounds composed of three key steps: global model distribution, local training, and global aggregation. At the beginning of each round \( t \), the central server distributes the current global model parameters \( w_t \) to a selected subset of \( K \) participating clients. Each client \( k \) then performs local training on its private dataset, yielding a locally optimized model \( w_t^k \). The local model update computed by client \( k \) is represented by:
\begin{equation}
    \Delta w_k =  w_t - w_t^k
\end{equation}

Here, \(\Delta w_k\) indicates the difference between the locally optimized parameters and the global model parameters received from the server at round \(t\). After completing local training, clients send these updates back to the central server, where the server aggregates using a weighted averaging method. Specifically, the global model parameters for the next iteration (\( w_{t+1} \)) are updated according to:
\begin{equation}
    w_{t+1} = w_t - \sum_{k=1}^{K} \frac{n_k}{n}\Delta w_k
\end{equation}

where \( n_k \) denotes the number of training samples at client \( k \), and \( n = \sum_{k=1}^{K} n_k \) is the total number of samples across all participating clients. This aggregation method effectively aggregates the local model update while preserving local data privacy and minimizing communication costs. Even though the Federated Averaging (FedAvg) algorithm performs well with homogeneous client data, it struggles under heterogeneous and non-IID conditions, which are common in EVCS due to variations in data distribution, usage, and infrastructure. To improve convergence in such settings, Federated Averaging with Server Momentum (FedAvgM) is introduced by Hsu et al.~\cite{hsu2019measuring}. In this approach, the server maintains a momentum buffer \( v_t \) that accumulates historical updates. The aggregated update from all clients is defined as:

\begin{equation}
\Delta w = \sum_{k=1}^{K} \frac{n_k}{n} (w_t - w_t^k),
\end{equation}

where $w_t$ is the global model at round $t$, $w_t^k$ is the model after local training on client $k$, $n_k$ is the number of samples on client $k$ and $n = \sum_{k=1}^{K} n_k$ is the total number of samples. The server momentum updates are then computed as:

\begin{equation}
v_{t+1} = \beta v_t + \Delta w,
\end{equation}

\begin{equation}
\qquad w_{t+1} = w_t - \eta v_{t+1},
\end{equation}

where \( \beta \in [0, 1) \) is the server momentum coefficient and \( \eta \) is the server learning rate. This learning rate helps stabilize the updates and ensures consistent contribution from the momentum term across communication rounds.

\begin{equation}
v_{t+1} = \beta v_t + (1 - \beta)\Delta w,
\label{eq:ema_momentum}
\end{equation}

Moreover, Sun et al.~\cite{sun2024role} 
adopt a server momentum variant inspired by the exponential moving average (EMA) to further smooth the global update and reduce instabilities during convergence. In this variant, the momentum update is modified as shown in Equation~\ref{eq:ema_momentum}, where \( \beta \) controls the influence of past updates and \( (1 - \beta) \) balances the effect of new updates. This ensures adaptive and smoother convergence over time, especially under non-IID settings, by giving more weight to the most recent updates while still incorporating the momentum from earlier updates. This technique leads to a more stable and efficient optimization process. In a special case, when $\beta = 0$ and $\eta = 1$, the update reduces to the standard Federated Averaging (FedAvg) algorithm, as no momentum is applied. This simplifies the process to incorporate only the most recent updates, eliminating any smoothing or momentum effect.

\section{Federated Learning Anomaly Detection System}

\begin{figure}[htbp]
    \centering
    \includegraphics[width=\columnwidth]{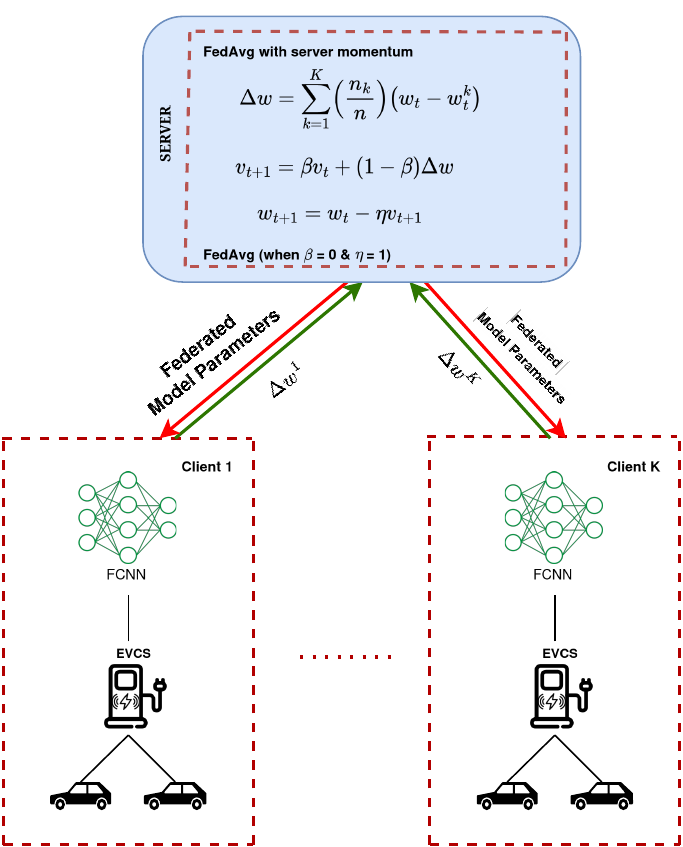}
    \caption{Federated Learning system architecture using FedAvg ($\beta = 0$, $\eta = 1$) and FedAvg with Server Momentum ($\beta > 0$).}
    \label{fig:FL_Model}
\end{figure}

In this research, a FL-based anomaly detection framework is proposed specifically for EVCS networks. Unlike generic FL applications, EVCS environments introduce unique operational challenges due to diverse hardware configurations, varying user charging behaviours, different charging protocols such as OCPP and ISO 15118, and exposure to multiple cyber-physical attack vectors. These factors contribute to both system heterogeneity and non-independent data distributions across charging stations, which directly affect anomaly detection performance.

The  FL architecture designed to address these challenges is illustrated in Figure~\ref{fig:FL_Model}. In this proposed model, each EVCS client independently trains a local model using its operational data over $E$ local epochs. Upon completion of local training, updated model parameters are transmitted to the central server, where aggregation is performed using either standard Federated Averaging (FedAvg) or Federated Averaging with Server Momentum (FedAvgM). FedAvgM incorporates a momentum buffer to enhance stability and convergence, particularly under heterogeneous and non-IID data conditions~\cite{hsu2019measuring}~\cite{sun2024role}. This iterative global training continues until model convergence while ensuring that raw operational data remains local to each EVCS, thereby preserving user and system privacy.

The detailed aggregation mechanism is described in Algorithm~\ref{alg:fedavgm}, where FedAvgM extends FedAvg by incorporating server momentum. When \(\beta = 0\) and \(\eta = 1\), FedAvgM reduces to the standard FedAvg. The algorithm is adapted from ~\cite{sun2024role}, incorporating additional details specific to our experiment.

\begin{algorithm}[]
\caption{Federated Averaging with Server Momentum (FedAvgM)}
\label{alg:fedavgm}
\textbf{Server executes:}
\begin{enumerate}[label=\arabic*:]\setcounter{enumi}{0}
    \item Initialize global model $w_0$ and server momentum buffer $v_0 = 0$
    \item \textbf{for} each round $t = 0, 1, 2, \dots$ \textbf{do}
    \item \quad Sample random set of $m$ clients $S_t$
    \item \quad \textbf{for} each client $k \in S_t$ \textbf{in parallel do}
    \item \quad \quad $w_t^k \gets$ \textsc{ClientUpdate}$(k, w_t)$
    \item \quad \textbf{end for}
    \item \quad Compute aggregated update: $\Delta w = \sum_{k=1}^K \frac{n_k}{n} (w_t - w_t^k)$
    \item \quad Update server momentum: $v_{t+1} = \beta v_t + (1 - \beta) \Delta w$
    \item \quad Update global model: $w_{t+1} = w_t - \eta v_{t+1}$
    \item \textbf{end for}
\end{enumerate}
\vspace{1em}
\textbf{ClientUpdate}$(k, w)$:
\begin{enumerate}[label=\arabic*:]\setcounter{enumi}{0}
    \item Split local data $P_k$ into batches $B$
    \item \textbf{for} each local epoch $i = 1$ to $E$ \textbf{do}
    \item \quad \textbf{for} each batch $b \in B$ \textbf{do}
    \item \quad \quad $w \gets w - \eta \nabla \ell(w; b)$
    \item \quad \textbf{end for}
    \item \textbf{end for}
    \item \textbf{return} $w$
\end{enumerate}
\end{algorithm}

To support effective anomaly detection, each EVCS station collects multi-modal operational data, which includes power consumption logs (charging current, voltage, delivered energy), session-based billing data (transaction ID, cost, duration), communication logs from OCPP protocol exchanges, and system-level logs (CPU usage, firmware version, process activity). This diverse data enables the detection of EVCS-specific attack scenarios such as billing fraud, power theft, firmware injection, denial-of-service attacks, communication protocol manipulation, and cryptojacking. By learning to distinguish between normal and malicious operating patterns across these modalities, the anomaly detection model identifies deviations indicative of malicious behaviour while maintaining full privacy for individual stations and users.

EVCS inherently exhibits substantial system-level heterogeneity across deployment sites. Variations in charging infrastructure, hardware specifications, and grid interaction patterns contribute to heterogeneous computational capabilities and non-uniform data distributions among clients~\cite{das2020electric}~\cite{ma2024exploring}. Such heterogeneity presents significant challenges for FL, particularly in ensuring stable model convergence and generalization. To address this, the anomaly detection framework utilizes a Feedforward Neural Network (FNN) architecture. Neural networks are well-suited for this task because they can capture the complex and non-linear relationships present in the structured tabular data derived from sources such as power delivery logs, communication records, and system diagnostics. While FNN provides competitive performance for static feature spaces, future work may explore integrating temporal models such as Long Short-Term Memory (LSTM) or Gated Recurrent Unit (GRU) to capture time-series dependencies in charging behaviours, power flows, and session dynamics. The proposed FL-based framework thus enables continuous collaborative model improvement while fully preserving privacy in distributed EVCS environments.

\section{Experimental Setup}

\subsection{Model Architecture and Training Configuration}

To evaluate the effectiveness of FL for anomaly detection in EVCS, a comparison was performed between a centralized model and a decentralized FedAvg approach, both utilizing the same Feedforward Neural Network (FNN) architecture. The centralized model was trained and evaluated using the entire dataset. For the FL setup, the FedAvg algorithm was employed by distributing the dataset among 10 clients, with each client performing 5 local epochs across 10 communication rounds. Across all experiments, a consistent Multilayer Perceptron (MLP) architecture was utilized as the base model. This architecture comprises two fully connected hidden layers with 64 neurons each and ReLU activation functions. To improve training stability and prevent overfitting, Batch Normalization and Dropout layers (dropout rate = 0.3) were applied after each hidden layer. A final sigmoid-activated output layer was used to perform binary classification. The model was compiled using the Adam optimizer with a learning rate of 0.001 and binary cross-entropy as the loss function, with accuracy as the evaluation metric. For FedAvgM, we consistently used a server learning rate of 1 following the configuration used by Hsu et al.~\cite{hsu2019measuring} to ensure consistency in experimental comparison.

\subsection{Dataset and Pre-processing}

\begin{figure}[htbp]
    \centering
    \includegraphics[width=\columnwidth]{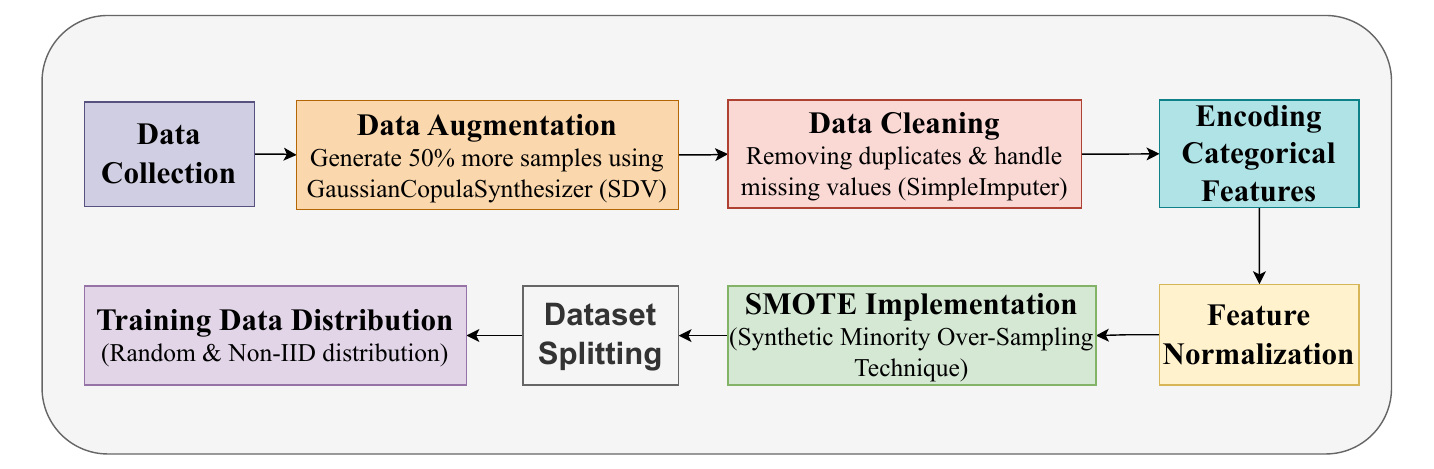}
    \caption{Data Preprocessing}
    \label{fig:data_preprocessing}
\end{figure}

The validation of anomaly detection systems (ADS) relies heavily on the dataset quality and diversity, ensuring both normal and anomalous scenarios are present. In this research, the CICEVSE2024 dataset \cite{CICEVSE2024} is used, which includes both benign and attack cases.  After data collection, a thorough preprocessing pipeline was applied to prepare the dataset for model training. As shown in Figure~\ref{fig:data_preprocessing}, preprocessing includes data augmentation, cleaning, encoding, and normalization. To increase data volume and variability, augmentation was performed using the GaussianCopulaSynthesizer from the SDV library, generating synthetic samples equal to 50\% of the original dataset. This step helped to reduce overfitting, improve generalization, and simulate realistic variations in both attack and benign scenarios. The dataset was then cleaned by removing duplicates and handling missing values using SimpleImputer. Categorical variables were encoded using One-Hot Encoding for ordinal features and Label Encoding for nominal features, followed by normalization with StandardScaler to ensure consistent feature scaling. To address class imbalance, particularly the underrepresentation of benign instances, the Synthetic Minority Over-sampling Technique (SMOTE) was applied. The dataset was then split into 80\% training and 20\% testing. Lastly, the training data was distributed across clients using two strategies: a random distribution and a non-IID distribution in which clients received varying proportions of benign and attack data to simulate realistic heterogeneity across EVCS nodes. The testing dataset remains the same for all clients. In this way, the dataset is prepared for the training phase of the anomaly detection models.

\subsection{Simulation of System Heterogeneity and Non-IID Conditions}

To simulate realistic FL scenarios, both system heterogeneity and non-IID data were simulated across clients. System heterogeneity was introduced by assigning different numbers of local training epochs to clients in a round-robin (circular) manner from the set \([1, 2, 3, 4, 5]\), reflecting the variability in computational capabilities and resource availability across EVCS. Instead of using a fixed number of local epochs for all clients, each client was trained for a different number of epochs based on its assigned value. This method also aligns with the approach used by Li et al.~\cite{li2020federated} to model system heterogeneity in FL. In addition to system-level variability, data heterogeneity was simulated by distributing varying proportions of benign and attack samples to each client. In this way, system heterogeneity and non-IID conditions were simulated to reflect the challenges of real-world federated EVCS environments closely.

The experiments were conducted using a machine running the Ubuntu operating system with a 64-bit x86\_64 architecture. The processor was an AMD EPYC 9334 32-Core CPU, with 8 vCPUs allocated for the experiment. The system was equipped with 32 GB of RAM. Virtualization was provided through VMware using full virtualization, and the system had a single NUMA node. The FL experiment was implemented using the Flower framework, and Jupyter Notebook was used as the interface to run the experiment.


\section{Experimental Results}

EVCS operate under diverse real-world conditions, including variations in power demand, user behavior, hardware configurations, and network latency. These operational differences lead to naturally non-IID and heterogeneous data distributions across stations, presenting significant challenges for anomaly detection. To evaluate the effectiveness of learning models in such settings, this section presents experiments comparing centralized and FL approaches, simulating realistic EVCS deployment scenarios. The goal is to assess the performance of decentralized training while addressing the privacy and system variability constraints inherent to EVCS infrastructures.


\begin{table}[htbp]
\caption{Performance comparison between centralized ML models and FL models (FedAvg and FedAvgM) under IID data at 10 communication rounds (values in \%).}
\begin{center}
\begin{tabular}{|c|c|c|c|c|}
\hline
\textbf{Model} & \textbf{Accuracy} & \textbf{Precision} & \textbf{Recall} & \textbf{F1 Score} \\
\hline
\makecell{Centralized\\approach} & 95.16 & 94.77 & 95.54 & 95.16 \\
FL (FedAvg) & 96.15 & \textbf{98.69} & 93.54 & 96.04 \\
FL (FedAvgM) & \textbf{96.61} & 97.24 & \textbf{97.11} & \textbf{97.17} \\
\hline
\end{tabular}
\label{tab:performance_comparison}
\end{center}
\end{table}



The accuracy, precision, recall, and F1 score of the centralized and decentralized FL models using the same FNN architecture are shown in Table~\ref{tab:performance_comparison}. This experiment aims is to demonstrate that even with data privacy, the decentralized FL-based models perform comparably or even better. As observed, the FedAvg and FedAvgM approaches achieved superior overall performance in terms of accuracy, precision, recall, and F1 Score compared to the centralized FNN model. Notably, the FedAvgM approach achieved the highest accuracy at 96.61\%, outperforming both the centralized model (95.16\%) and the standard FedAvg (96.15\%). This improvement in accuracy, coupled with its superior recall (97.11\%) and F1 Score (97.17\%), highlights FedAvgM’s enhanced capability in correctly identifying anomalies while maintaining balanced classification performance. Among the two FL approaches, FedAvgM achieved higher accuracy, recall, and F1 Score than FedAvg, further indicating its robustness under the given conditions. This FL approach not only enhances model generalization across diverse datasets but also ensures data privacy and security, which are critical in EVCS environments. The results demonstrate the potential of decentralized anomaly detection using FL, offering a compelling alternative to traditional centralized methods.

\begin{figure}[htbp]
    \centering
    \includegraphics[width=\columnwidth]{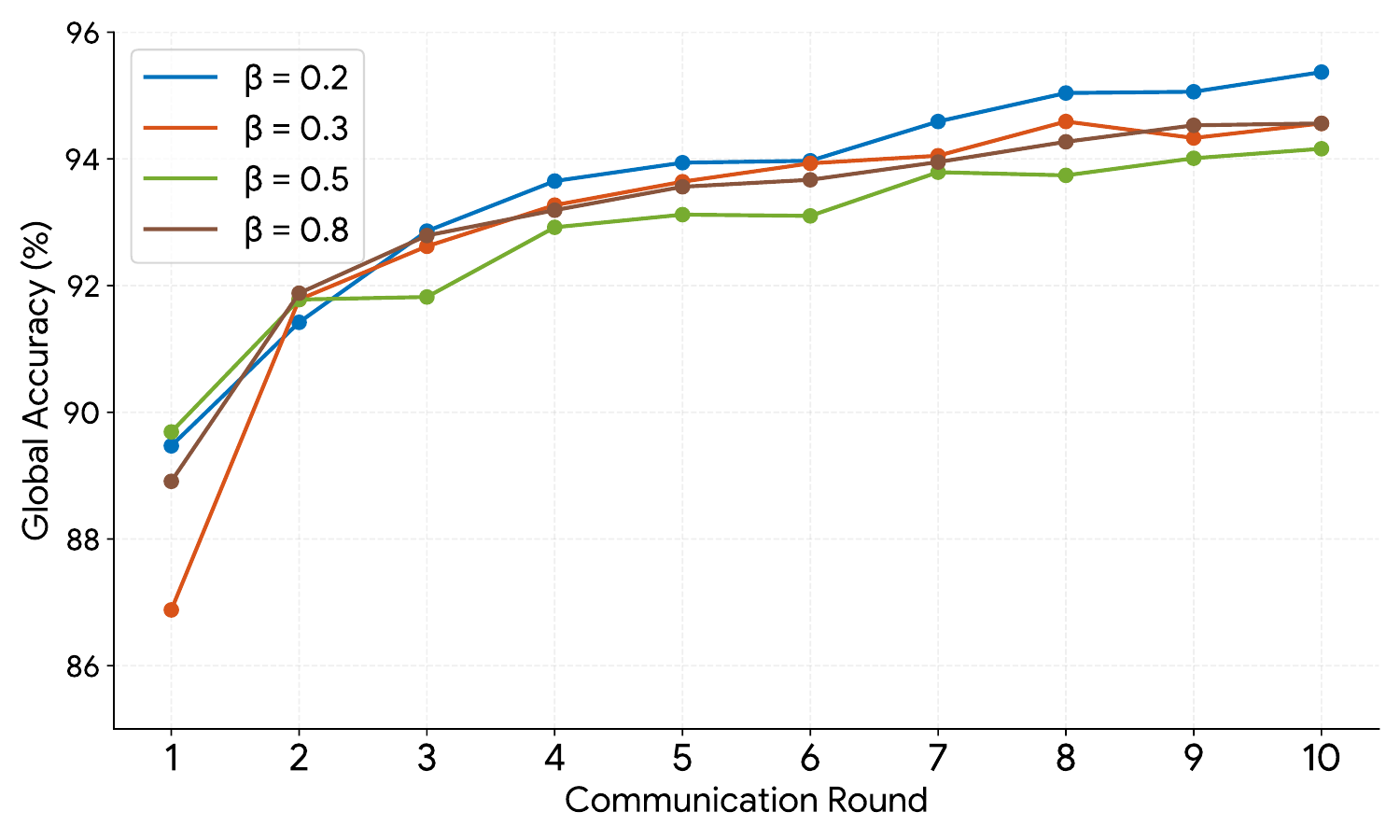}
    \caption{FedAvgM performance across server momentum values (\(\beta\)) with 10 clients and 5 local epochs.}
    \label{fig:momentum_plot}
\end{figure}

Further experiments were performed to investigate FedAvg under the conditions of system and data heterogeneity.  After introducing these heterogeneity conditions, the performance of FL (FedAvg) declined, and convergence occurred at a slower rate compared to the homogeneous setting. To address these challenges, FedAvgM was explored due to its enhanced capability to handle such heterogeneity. Parameter tuning was performed across various momentum values (\(\beta \in \{0.1, 0.2, 0.3, 0.4, 0.5, 0.6, 0.7, 0.8, 0.9\}\)) to determine the optimal server momentum. As shown in Figure~\ref{fig:momentum_plot}, only selected momentum values are presented for clarity. A momentum value of \(\beta = 0.2\) consistently yielded the best performance and was therefore used in all subsequent FedAvgM experiments.

\begin{table}[htbp]
\caption{Performance comparison of FedAvg (IID), FedAvg (Non-IID), and FedAvgM (Non-IID) with varying no. of clients after 10 communication rounds. } 
\begin{center}
\begin{tabular}{|c|c|c|c|}
\hline
\textbf{Model} & \textbf{Clients} & \textbf{Accuracy} & \textbf{F1 Score} \\
\hline
\multirow{3}{*}{FedAvg (IID)} & 10 & 96.15 & 96.04 \\
                              & 15 & 95.42 & 95.22 \\
                              & 20 & 94.83 & 94.69 \\
\hline
\hline
\multirow{3}{*}{FedAvg (Non-IID)} & 10 & 93.63 & 91.28 \\
                                  & 15 & 93.56 & 91.11 \\
                                  & 20 & 92.77 & 91.03 \\
\hline
\multirow{3}{*}{FedAvgM (Non-IID)} & 10 & \textbf{94.65} & \textbf{92.24} \\
                                   & 15 & \textbf{94.11} & \textbf{92.06} \\
                                   & 20 & \textbf{93.93} & \textbf{91.82} \\
\hline
\end{tabular}
\label{tab:fedavgm_results}
\end{center}
\end{table}

\begin{figure}[htbp]
    \centering
    \includegraphics[width=0.48\columnwidth]{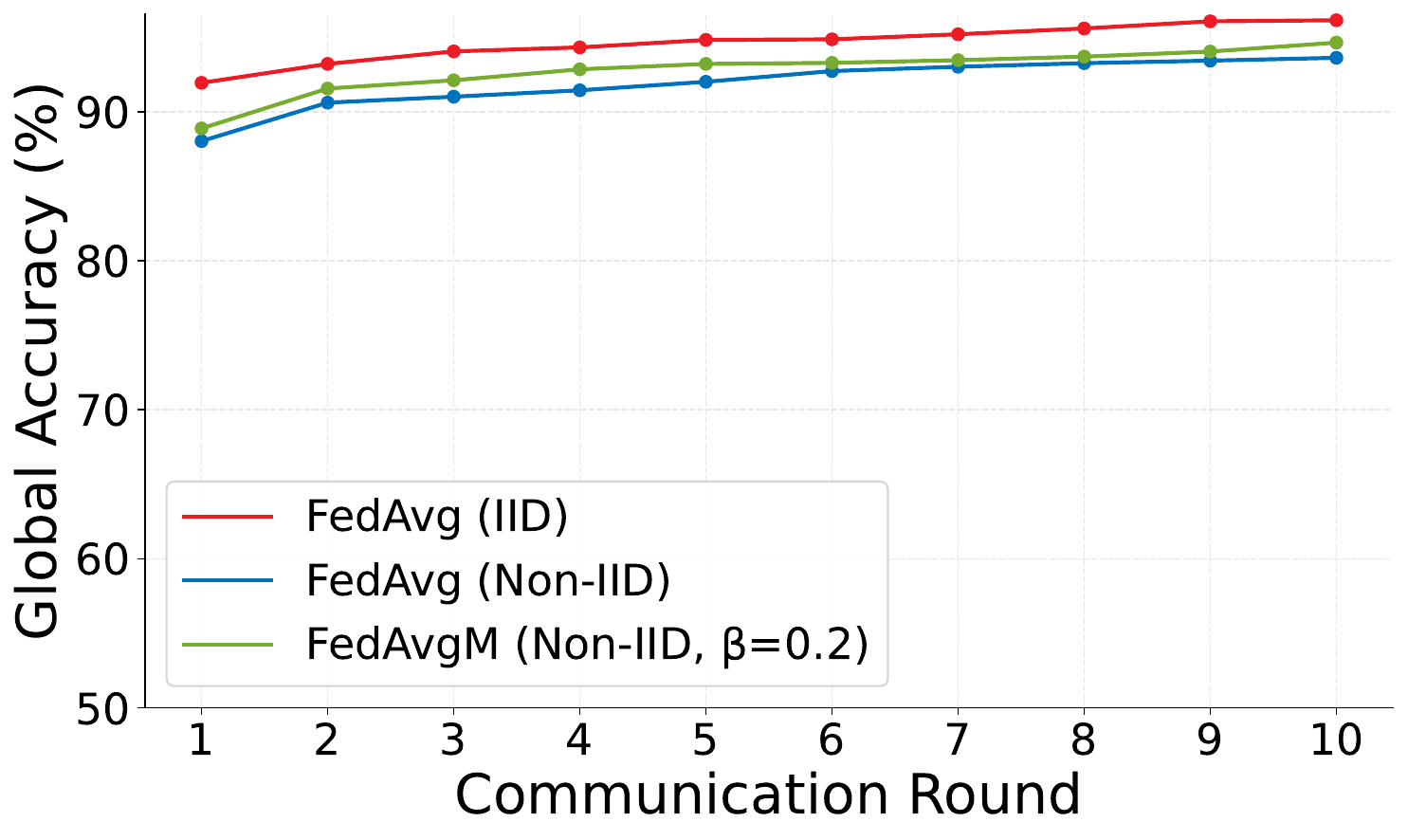}
    \hfill
    \includegraphics[width=0.48\columnwidth]{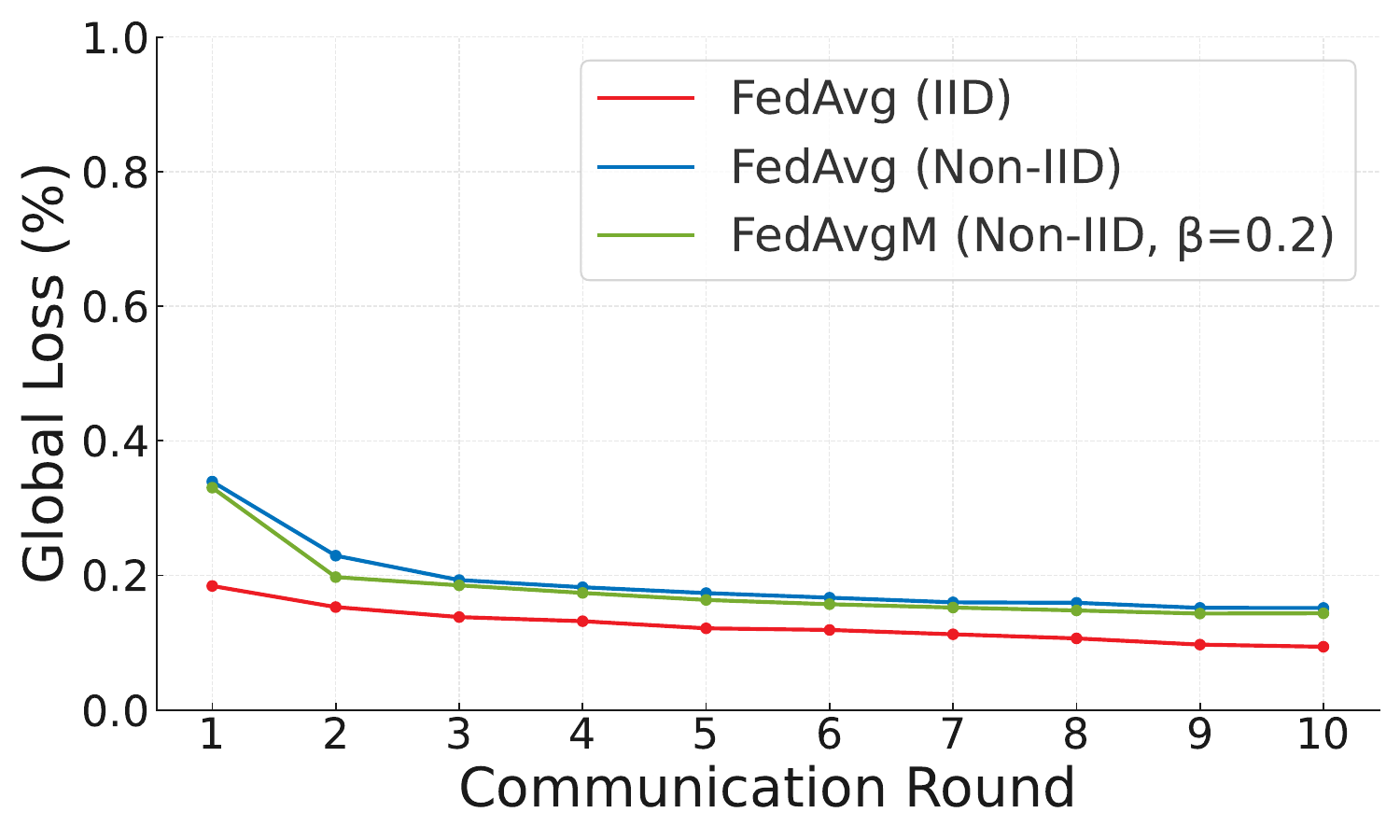}
    \caption{Comparison of FedAvg under ideal IID conditions vs. FedAvg \& FedAvgM under system and data heterogeneity, showing model accuracy \& loss across communication rounds.}
    \label{fig:fedavg_methods}
\end{figure}

The Table~\ref{tab:fedavgm_results} presents the experimental results obtained after 10 communication rounds, comparing the accuracy and F1 Score of FedAvg and FedAvgM across different client configurations.
While FedAvg achieves the highest performance under IID conditions, its accuracy and F1 Score decline noticeably when exposed to system and data heterogeneity. In contrast, FedAvgM demonstrates improved robustness, consistently outperforming FedAvg under the same heterogeneous settings.
For instance, with 15 clients, FedAvg under heterogeneity achieves an accuracy of 93.56\% and an F1 Score of 91.11\%, whereas FedAvgM improved this to 94.11\% and 92.06\% respectively. A similar trend is observed for 10 and 20 clients, where FedAvgM not only reduces the performance gap caused by heterogeneity but also brings the metrics closer to those observed under ideal IID conditions. As shown in Figure~\ref{fig:fedavg_methods}, FedAvgM also achieves faster convergence and lower loss compared to FedAvg under system and data heterogeneity. While FedAvg performs well under IID settings, its performance declines in heterogeneous scenarios. In contrast, FedAvgM consistently shows better optimization, with similar trends observed across 10, 15, and 20 client setups.

\section{Discussion}

The experimental results clearly demonstrate that the FedAvgM algorithm offers enhanced robustness and stability when applied to real-world EVCS environments characterized by diverse operational heterogeneity. In actual EVCS deployments, heterogeneity arises from several domain-specific factors, such as differences in charger type (e.g., Level 2 AC chargers vs DC fast chargers), customer usage patterns (public commercial stations with high transactional volumes vs. private residential stations with fewer sessions), and grid integration (e.g., stations participating in Vehicle-to-Grid (V2G) energy exchange vs. standalone chargers)~\cite{das2020electric}~\cite{ma2024exploring}. Such operational variability directly results in non-IID data distributions, system resource variability, and asynchronous client updates during FL. Importantly, the impact of accurate anomaly detection in EVCS extends beyond conventional classification performance metrics. Failure to detect attacks can lead to substantial financial losses through fraudulent billing, energy theft, or cryptojacking, which drains local energy resources. Additionally, undetected anomalies may cause operational disruptions in EV charging operations, compromise grid stability during V2G interactions, or even damage expensive charger hardware due to intentional power overloads or firmware tampering. FL enables collaborative model training across distributed EVCS stations while avoiding centralized data aggregation, which is a particularly important advantage given the sensitivity of EVCS customer data, such as charging behavior, personal location, and billing records.

By adopting an FL framework, this study demonstrates the potential of decentralized model training to achieve stable global convergence under severe non-IID and heterogeneous conditions that reflect the operational variability of EVCS networks. This study underscores the strength of FL in addressing system and data heterogeneity in real-world deployments.  FedAvgM, among the aggregation strategies explored, shows how sufficiently well-designed algorithmic choices can further enhance convergence stability, even within these environments. Overall, the proposed FL-based anomaly detection system offers a scalable, privacy-preserving, and operationally resilient solution to the cybersecurity and data privacy challenges associated with modern electric vehicle charging infrastructures.

\section{Future Work}
While the proposed framework demonstrates strong potential, addressing the growing complexity and operational diversity of real-world EVCS deployments requires further investigation. Future work could examine more advanced aggregation strategies such as Scaffold~\cite{karimireddy2020scaffold}, FedNova~\cite{wang2020tackling}, and FedOpt~\cite{reddi2020adaptive} which are specifically designed to mitigate client drift, resource imbalance, and asynchronous updates, challenges frequently encountered in EVCS networks with heterogeneous charging infrastructure, diverse customer usage patterns, and varying levels of V2G participation.
With the rapid adoption of EVs, EVCS datasets are also expanding in both scale and complexity, encompassing not only routine operational data but also a rising incidence of sophisticated cyber-physical threats. This evolution creates new opportunities to develop advanced neural architectures capable of identifying subtle behavioral anomalies, while enabling rigorous evaluations of FL performance under realistic and adversarial conditions.

\section{Conclusion}
In this paper, we present an empirical study addressing system and data heterogeneity in federated learning (FL) for anomaly detection within Electric Vehicle Charging Station (EVCS) networks. The evaluation compares a centralized approach with decentralized FedAvg and FedAvgM under both IID and heterogeneous conditions, simulating realistic operational diversity across EVCS clients, including variations in charger type, usage patterns, and grid integration. While FedAvg exhibited performance degradation under non-IID scenarios, FedAvgM demonstrated that well-chosen aggregation algorithms can enhance resilience and accuracy across diverse client configurations. A comparative assessment with the centralized approach, using the same neural network architecture, confirmed that FL can achieve comparable or superior anomaly detection performance while preserving data privacy, which is critical given the sensitivity of EVCS billing, customer usage, and location data. As EVCS operate as critical IoT nodes within broader smart energy and transportation infrastructures, our findings underscore the potential of FL-based anomaly detection to strengthen security across large-scale IoT deployments. Although FedAvgM improved robustness, convergence instability remains a challenge under extreme heterogeneity. Overall, FedAvgM offers substantial promise for enhancing the robustness, privacy, and scalability of anomaly detection in IoT-enabled EVCS infrastructures. Future work could explore larger-scale experiments with more diverse real-world datasets, advanced aggregation algorithms such as Scaffold, FedOpt, or adaptive client selection, along with comprehensive evaluations of FL-based anomaly detection in IoT environments, particularly under adversarial or extreme operational conditions.

\end{document}